\documentclass{article}

\usepackage{arxiv}

\usepackage[utf8]{inputenc} % allow utf-8 input
\usepackage[T1]{fontenc}    % use 8-bit T1 fonts
\usepackage[hidelinks]{hyperref}       % hyperlinks
\usepackage{url}            % simple URL typesetting
\usepackage{booktabs}       % professional-quality tables
\usepackage{amsmath,amssymb,amsfonts}%
\usepackage{nicefrac}       % compact symbols for 1/2, etc.
\usepackage{microtype}      % microtypography
\usepackage{cleveref}       % smart cross-referencing
\usepackage{lipsum}         % Can be removed after putting your text content
\usepackage{graphicx}
\usepackage{natbib}
\usepackage{doi}
\usepackage{multirow}
\usepackage{xcolor}
\hypersetup{
    colorlinks,
    linkcolor={red!50!black},
    citecolor={blue!50!black},
    urlcolor={black}
}
\title{Graph Spring Neural ODEs for Link Sign Prediction}

\newif\ifuniqueAffiliation
\uniqueAffiliationtrue

\ifuniqueAffiliation % Standard variant of author block
\author{Andrin Rehmann\\
	Pasteur Labs\\
	New York\\
	USA\\
	\texttt{andrin.rehmann@simulation.science} \\
	\And
	\href{https://orcid.org/0000-0003-3937-3704}{\includegraphics[scale=0.06]{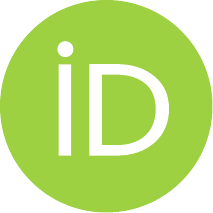}\hspace{1mm}Alexandre Bovet} \\
	Department of Mathematical Modeling and Machine Learning\\
	Digital Society Initiative\\
	University of Zurich\\
	Switzerland\\
	\texttt{alexandre.bovet@uzh.ch} \\
}
\else
\usepackage{authblk}

\newbox{\orcid}\sbox{\orcid}{\includegraphics[scale=0.06]{orcid.pdf}} 
\author[1]{%
	\href{https://orcid.org/0000-0000-0000-0000}{\usebox{\orcid}\hspace{1mm}David S.~Hippocampus\thanks{\texttt{hippo@cs.cranberry-lemon.edu}}}%
}
\author[1,2]{%
	\href{https://orcid.org/0000-0000-0000-0000}{\usebox{\orcid}\hspace{1mm}Elias D.~Striatum\thanks{\texttt{stariate@ee.mount-sheikh.edu}}}%
}
\affil[1]{Department of Computer Science, Cranberry-Lemon University, Pittsburgh, PA 15213}
\affil[2]{Department of Electrical Engineering, Mount-Sheikh University, Santa Narimana, Levand}
\fi

\renewcommand{\headeright}{}
\renewcommand{\undertitle}{}
\renewcommand{\shorttitle}{Graph Spring Neural ODEs for Link Sign Prediction}

\hypersetup{
pdftitle={Graph Spring Neural ODEs for Link Sign Prediction},
pdfsubject={cs.ML},
pdfauthor={Andrin Rhemann, Alexandre Bovet},
pdfkeywords={Graph machine learning, Signed Graphs, Node Representation Learning, Sign Prediction, Graph Neural Ordinary Differential Equations},
}

\begin{document}
\maketitle

\begin{abstract}
Signed graphs allow for encoding positive and negative relations between nodes and are used to model various online activities. Node representation learning for signed graphs is a well-studied task with important applications such as sign prediction. While the size of datasets is ever-increasing, recent methods often sacrifice scalability for accuracy. We propose a novel message-passing layer architecture called Graph Spring Network (GSN) modeled after spring forces. We combine it with a Graph Neural Ordinary Differential Equations (ODEs) formalism to optimize the system dynamics in embedding space to solve a downstream prediction task. Once the dynamics is learned, embedding generation for novel datasets is done by solving the ODEs in time using a numerical integration scheme.
Our GSN layer leverages the fast-to-compute edge vector directions and learnable scalar functions that only depend on nodes' distances in latent space to compute the nodes' positions.
Conversely, Graph Convolution and Graph Attention Network layers rely on learnable vector functions that require the full positions of input nodes in latent space.
We propose a specific implementation called Spring-Neural-Network (SPR-NN) using a set of small neural networks mimicking attracting and repulsing spring forces that we train for link sign prediction. Experiments show that our method achieves accuracy close to the state-of-the-art methods with node generation time speedup factors of up to 28,000 on large graphs.
\end{abstract}

% keywords can be removed

\section{Introduction}\label{sec1}

Signed graphs distinguish negative and positive edges, also called link signs. They are crucial to modeling relations in online platforms, such as attitudes between users in social media or preferences between users and products in online stores.
The ability to predict new links and their type - either negative or positive - between users is crucial for analyzing and understanding dynamics in social networks \citep{leskovec2010signed} or for improving recommender systems \citep{huang2023negative}.

Methods for link sign prediction often involve social balance theory \citep{cartwright1956structural}, which postulates that triangles in a network, where edges represent attitudes or relations of liking, tend to be balanced, i.e., the product of the signs of three edges in a triad is positive. However, not all triadic relationships in real-world networks satisfy this rule, and some edges do not participate in a triad. Hence, reliable predictions require extending or replacing rule-based approaches with heuristic models. Typically, the problem is solved in a two-step procedure. First, a multidimensional vector, or node embedding, is generated for all nodes based on the available edge signs and graph structure. Subsequently, the Euclidean distance between the node embeddings of adjacent nodes serves as an input to a logistic regression model, which is then used to infer the signs of the edges \citep{huang2021sdgnn, derr2018signed}. The embedding vectors are frequently generated using a graph neural network (GNN), which iteratively improves the embedding through training. Often, a term capturing social balance theory is included in the loss function. Although architecture designs vary across different methods, the overall framework remains similar. The approach offers high inference accuracy; however, the process does not scale well for large graphs.

In this work, we generate the embeddings from the time-evolved states obtained by solving second-order ordinary differential equations (ODEs) involving the nodes' positions. To this end, each node is associated with a vector position in the embedding space, which is initially random. By cleverly designing a function for the 2nd-order time derivative of the node positions, the system evolves forward in time, leading to final node positions that can be leveraged to solve the downstream link sign prediction task. We model the dynamics of the second-order ODEs with two learnable scalar functions. The first scalar function is evaluated on each edge and reads edge and adjacent node attributes. Its resulting value is then used to scale the direction vector associated with the edge in latent space, resulting in an edge force vector. Those vectors are then summed up and scaled with the second learnable function, constituting the position vector's second-order derivative in latent space. The ODEs system is then solved with a numerical integrator, whereas the final node positions in latent space are taken as the node embeddings. The continuous modeling approach using learnable functions to model the time derivative of a system's state corresponds to the framework of Graph Neural ODE's \citep{poli2019graph,rusch2022graph,sun2024dynamise}.
Whereas previous research leveraged common neighborhood coupling functions such as Graph Convolutional Networks (CCN) or Graph Attention Networks (GAN), we propose a novel message-passing layer architecture termed Graph Spring Network (GSN). In contrast to GCN or GAN, the number of learnable parameters of the GSN layer is independent of the number of dimensions of the latent space. Subsequently, the computational time of inference and training is lower due to the reduced parameter count.
In contrast to discrete modeling approaches used to solve link sign prediction task \citep{liu2022lightsgcn, huang2021sdgnn, li2020learning, islam2018signet, ren2024link, chung2023learning}, the continuous modeling approach allows us to learn the dynamics of the ODE once, and generate node embeddings on unseen datasets without training. This can be done because the solution is found by solving the ODEs with a numerical solver, which gives our approach good generalization properties.

We propose two distinct approaches to model the GSN layer, Spring (SPR) and Spring Neural Network ({SPR-NN}). {SPR} is based on Hooke's law - an equation to model the dynamics in physical springs and {SPR-NN} leverages a small, shallow artificial neural network. The constants specifying the dynamics of spring equations in {SPR} and the neural network parameters in {SPR-NN} are optimized using a gradient-based optimization method. Both methods achieve competitive accuracy scores across all examined datasets, where {SPR-NN} is significantly more accurate than {SPR}. At the same time, our method demonstrates runtime improvements by a factor $10^5$ for large graphs compared to a methods with similar accuracy. The main contributions of this paper are as follows:

\begin{itemize}
    \item We propose a novel message passing architecture, Graph Spring Network (GSN), for scalable node embedding generation and derive its computational complexity.
    \item We propose two novel methods for link sign prediction, Spring (SPR) and Spring Neural Network ({SPR-NN}), based on GSN.
    \item We demonstrate their effectiveness and performance on real-world signed graphs, showing that SPR-NN achieves comparable accuracy to state-of-the-art methods while outperforming them in terms of runtime.
\end{itemize}

\section{Related Work}
\label{related}

\subsection{Link Sign Prediction with Discrete Modeling}

In discrete modeling approaches, a neural network with a fixed number of layers is used to predict the next state of interest of the inference. The first methods proposed for sign inference in signed graphs are based on formulating rules-based extensions of the structural balance theory \citep{guha2004propagation,song2015link}. \cite{leskovec2009community} proposed a logistic regression trained on several extracted node features, such as degrees and the number of incoming negative and positive edges. Advancements in machine learning and increased computing powers enabled graph embedding generation approaches, pioneered by {Deepwalk} \citep{perozzi2014deepwalk} and {Node2Vec} \citep{grover2016node2vec}. Both methods train and generate node embedding using stochastic gradient descent. The node vectors are generated so that spatial proximity in the embedding space correlates with proximity on the structural level in the graph. However, these approaches target link prediction for unsigned networks. Later methods were explicitly designed for link sign prediction \citep{chung2023learning, ren2024link, shi2024structural}. For example, \cite{wang2017signed} introduced \textit{SiNE}, an embedding generation method integrating social balance theory for link sign prediction in social media. Signed Graph Convolutional Networks (\textit{SGCN}) is the first method to apply a graph convolutional network architecture specifically to undirected link sign prediction \citep{derr2018signed}. The approach has been refined with \textit{LightSGCN} \citep{liu2022lightsgcn}. Graph Convolutional Network architectures for directed link sign prediction were also proposed with \textit{BRIDGE} \citep{chen2018bridge}, \textit{SIGNET} \citep{islam2018signet}, \textit{SDGNN} \citep{huang2021sdgnn} and \textit{SELO} \citep{fang2023signed}. The attention mechanism was introduced to graph neural networks \citep{velivckovic2017graph}, which was then adapted to signed networks by \cite{huang2019signed} with \textit{SiGAT}  and refined and improved in follow-up work \citep{li2020learning, shi2024structural}.
A spectral graph theory and graph signal processing approach was also recently proposed by \cite{li2023signed} with \textit{SLGNN}. Other recent developments in the field include the study of adversarial attacks against link prediction in signed networks \citep{lizurej2023manipulating, lee2024spear} and link sign prediction in multilayer signed graphs \citep{ren2024link}.

\subsection{Continuous Modeling for Node Representation Learning}

In contrast to discrete modeling approaches, continuous modeling approaches learn the temporal dynamics of a system. Learning with continuous systems can be more memory efficient and less prone to over-smoothing when information propagation across a large graph is crucial \citep{xhonneux2020continuous}. \cite{chen2018neural} introduced the Neural ODEs framework as a family of continuous-depth models where the dynamics of ODEs is learned by modeling the first-order derivative of the state using an artificial neural network. The Graph Neural ODEs framework, extending {Neural ODEs} to graph-structured datasets, proposed by \cite{poli2019graph} defines the derivatives of nodes' representations as a combination of the current node representations,
the representations of neighbors, and the initial
values of the nodes. The approach was extended with a self-attention mechanism and an encoder network \citep{huang2020learning}. With \textit{GraphCon} \citep{rusch2022graph}, the Graph Neural ODE framework was extended by learning the second instead of first-order ODEs dynamics and adding a damping term for stability. Independently \cite{sanchez2019hamiltonian} proposed a GNN approach with learnable ODEs based on a Hamiltonian mechanics formalism, which was later extended by Kang \textit{et al.} \citep{kang2023node}. The Graph Neural ODE-based approach has recently been applied to link sign prediction in dynamic graphs with DynamicSE \citep{sun2024dynamise}.
Graph Neural ODEs also take inspiration from diffusion processes on graphs, such as random walks \citep{xhonneux2020continuous,chamberlain2021grand}. Signed graph diffusion approaches have also been applied to generate node representations of signed graphs \citep{jung2020signed,choi2023gread} and for link sign prediction with \textit{SDGNet} \citep{jung2020signed}.

\subsection{Force Directed Graph Algorithms}

Force-directed approaches model nodes as particles and evolve their positions forward in time with a numerical solver. Typically, they do it in two dimensions to visualize the final node positions with a scatter plot. \cite{fruchterman1991graph} proposed attracting spring forces between adjacent nodes and repulsing electrical force between all nodes. The combination ensures the closeness of connected nodes while retaining visual distinctiveness between nodes. This principle has been iteratively improved in follow-up work \citep{kamada1989algorithm, hadany2001multi}. Moving beyond visualization, force-directed approaches have also been applied to link prediction tasks \citep{rahman2020force2vec, lotfalizadeh2023force}, which can be done by increasing the embedding dimension and interpreting the node positions as node embeddings.

\section{Background}
\label{background}

\subsection{Problem Statement}
\label{problem}

We consider a signed undirected graph $G = (V, E, \sigma)$, with $V$ representing the set of nodes and $E$ denoting the set of edges and the function $\sigma: V \times V \rightarrow \{-1, 1\}$ assigning a sign of either -1 or 1 to each edge $(i, j) \in E$. The total number of nodes in the graph is expressed as $|V| = N$. We derive a secondary graph $G'$ from the original graph $G$ to evaluate our prediction method. In $G'$, a specified percentage $p_\text{hidden}$ of edge signs is obscured. The altered graph is defined as $G' = (V, E, \sigma') $, where $V$ and $E$ remain unchanged, while $\sigma': V \times V \rightarrow \{-1, 0, 1\}$ is designed as

\begin{equation}
\sigma^{\prime}(i, j) = \begin{cases}
\sigma(i, j) & \text{if } \mathbb{P}(Y > p_\text{hidden}) \\
0 & \text{otherwise},
\end{cases}
\end{equation}

where $Y$ is a random variable with a uniform distribution between 0 and 1 and $\mathbb{P}$ denotes the probability function. We aim to infer a sign prediction function $\widehat{\sigma}: V \times V \rightarrow \{-1, 1\}$, capable of approximating the original sign function $\sigma$ as closely as possible. Note that our approach allows us to set neutral values to the edges whose sign we wish to infer.
Our problem formulation differs from other formulations where edges in the test set are completely removed from the graph instead of marking them as neutral.

\section{Framework}
\label{SpringBasedForces}

\subsection{Graph Spring Network Layer}

Without any loss of generality, a message passing layer can be defined as \citep{bronstein2021geometric}
\begin{equation}
    \label{eq:mpl}
    \text{mpl}(\mathbf{x}_i) = \phi\left(\mathbf{y}_i, \mathbf{x}_i \bigoplus_{j \in N_i} \psi \bigl(\mathbf{x}_i, \mathbf{x}_j, \mathbf{z}_{i,j} \bigr)\right),
\end{equation}
where $N_i$ denotes the set of neighboring nodes of $i$, $\mathbf{z}_{ij}$ and $\mathbf{y_i}$ the static edge and node features, $\bigoplus$ is an aggregation operations, $\phi$ and $\psi$ differentiable functions and $\mathbf{x}_i\in \mathbb{R}^k$ the dynamic node representations. We define a specific message-passing layer called Graph Spring Network (GSN) as
\begin{equation}
    \label{eq:gsn}
    \text{gsn}(\mathbf{x}_i) = g(\mathbf{y}_i) \cdot \sum_{j \in N_i} f \bigl(\mathbf{z}_{i,j}, d(\mathbf{x}_{i}, \mathbf{x}_{j})\bigr) \cdot \frac{\mathbf{x}_j - \mathbf{x}_i}{d(\mathbf{x}_i, \mathbf{x}_j)}
\end{equation}
where $d$ denotes the Euclidean distance between two nodes
\begin{equation}
    \label{eq:d}
    d(\mathbf{x}_i, \mathbf{x}_j) = \sqrt{(\mathbf{x}_i - \mathbf{x}_j)^2}
\end{equation} and $f$, $g$ are differentiable scalar functions, which could be functions with learnable parameters such as a neural network. The
node features, $\mathbf{y}_i$, and edge features, $\mathbf{z}_i$, are both independent of the node positions in the embedding space. In contrast to other layers, such as a GCN or GAT, which use learnable vector functions with the full nodes' positions as input,
the learnable scalar functions involved with GSN only read the distance vectors between adjacent nodes and static graph features. Hence, the complexity of the learnable functions can be greatly reduced as the number of data processed is significantly lower. As detailed in section \ref{spring-forces}, we can construct two scalar functions that rely on a total of seven learnable parameters and still solve a downstream prediction task using the GSN layer.

\subsection{Graph Spring Network Layer Complexity}

We denote the size of the embedding space as $k$, $N$ as the number of nodes, and $M$ as the number of edges. The time complexity of a single graph convolution or a graph attention layer is $Mk + Nk^2$ \citep{blakely2021time}. We analyze the complexity of the GSN layer by decomposing Equation \ref{eq:gsn} into three operations
\begin{enumerate}
    \item Edge level transformation: \begin{equation}
        \mathbf{a}_{i,j} = f \bigl(\mathbf{z}_{i,j}, d(\mathbf{x}_{i}, \mathbf{x}_{j})\bigr) \cdot \frac{\mathbf{x}_j - \mathbf{x}_i}{d(\mathbf{x}_i, \mathbf{x}_j)}.
    \end{equation}
    \item Neighborhood aggregation: \begin{equation}
        \mathbf{b}_i = \sum_{j \in N_i} \mathbf{a}_{i,j}.
    \end{equation}
    \item Node level transformation: \begin{equation}
        g(\mathbf{y}_i) \cdot \mathbf{b}_i.
    \end{equation}
\end{enumerate}
Part 1 is made of a vector subtraction ($O(k)$), a computation of a vector distance ($O(1)$), and an evaluation of the function $f$ which is independent of $k$ ($O(1)$). Part 1 is done for all edges; hence, its complexity is $O(Mk)$. Part 2 is a summation of the $k$ dimensional vectors resulting from part 1. The operation is done for the average number of degrees ($O({\frac{M}{N}} k)$), which is repeated for all nodes; hence its complexity is $O(N\frac{M}{N} k + N k) = O(M k + N k)$. Part 3 is a vector scaling repeated for all nodes resulting in $O(Nk).$ Adding all parts together yields $O(Mk + Mk + Nk + Nk) = O(Mk + Nk)$.

\subsection{Temporal Dynamics}

We use a single GSN layer in combination with Graph Neural Ordinary Differential Equations \citep{poli2019graph}, obtaining the initial-value problem defined by
\begin{equation}
\label{eq:c1}
    \frac{d \mathbf{v}_i(t)}{dt} = \text{gsn}(\mathbf{x}_i(t)),
\end{equation}
\begin{equation}
\label{eq:c2}
    \frac{d \mathbf{x}_i(t)}{dt} = \mathbf{v}_i,
\end{equation}
\begin{equation}
\label{eq:c3}
    \mathbf{x}_i(0) \sim \mathcal{U}_{(-1, 1)}
\end{equation} and
\begin{equation}
\label{eq:c4}
    \mathbf{v}_i(t) = 0.
\end{equation}
which can be solved using any numerical ODE solver. Here $\mathbf{v}_i$ denotes the velocity of node $i$ and $\mathcal{U}_{(-1, 1)}$ the random uniform distribution on $(-1, 1)$.

\subsection{{SPR}: Spring Based Forces}
\label{spring-forces}

The central premise of our approach is that by customizing the scalar functions $f$ and $g$ and choosing appropriate parameters, we can influence the lengths of edges with unknown signs to either expand or contract to an appropriate length with respect to their surrounding. To this end, our first method uses an adapted version of Hooke's law, which describes the physical forces in a spring. A spring exerts a force linearly proportional to the difference between its resting length and actual length. For each edge type, positive, neutral, and negative, we assign distinct resting lengths: $l^{+}$ for positive, $l^{\pm}$ for neutral, and $l^{-}$ for negative edges. Accompanying these resting lengths are the respective stiffness coefficients, $\alpha^{+}$ for positive, $\alpha^{\pm}$ for neutral, and $\alpha^{-}$ for negative interactions, with all coefficients being real-valued parameters. We organize all learnable parameters in a vector $\theta$ . The scalar function $f$ is defined as
\begin{equation}
    f( \theta_f, \mathbf{z}_{ij}) =
    \begin{cases}
         \alpha^{\mp} \cdot \left(d(\mathbf{x}_j, \mathbf{x}_i) - l^{\mp} \right) &
         \text{if  $\sigma^{\prime}(i, j)$ is 0} \\
         \alpha^{+} \cdot \max\left(d(\mathbf{x}_j, \mathbf{x}_i) - l^{+}, 0 \right) &
         \text{if  $\sigma^{\prime}(i, j)$ is 1} \\
         -\alpha^{-} \cdot \max\left(l^{-} - d(\mathbf{x}_j, \mathbf{x}_i), 0\right) &
         \text{if $\sigma^{\prime}(i, j)$ is -1} \\
    \end{cases}
\label{eq:spring_scalarf}
\end{equation}
where
\begin{equation}
    \mathbf{z_{ij}} = \begin{bmatrix}
        d(\mathbf{x}_i, \mathbf{x}_j)
    \end{bmatrix}
\end{equation} and
\begin{equation}
    \theta_f = \begin{bmatrix}
        l^{+} &
        l^{\mp} &
        l^{-} &
        \alpha^{+} &
        \alpha^{\mp} &
        \alpha^{-} &
    \end{bmatrix}^{\top}.
\end{equation}

In the case of $\sigma^{\prime}(i, j) = 0$, equation \ref{eq:spring_scalarf} directly follows Hooke's law. If the edge is negative ($\sigma^{\prime}(i, j) = -1$), we apply a force that pushes the adjacent nodes apart from each other. If two nodes lie at a distance greater than $l^{-}$ no force is applied, hence they can freely drift apart from each other. Analogously, for positive edges ($\sigma^{\prime}(i, j) = 1$), we apply an attraction force between connected nodes only when they are farther apart than $l^{+}$. If they are closer, no forces are applied. We define function $g$ as
\begin{equation}
    g(\beta, \mathbf{y}_i) = \left(\min(1, \text{deg}_i / P_{80}) \cdot \beta + 1\right)
    \label{eq:spring_g}
\end{equation}
where
\begin{equation}
    \mathbf{y}_i = \begin{bmatrix}
        \text{deg}_i, p_{80}
    \end{bmatrix}.
\end{equation}
The function $g$ ensures stronger forces for nodes with a higher connectivity. The term $p_{80}$ represents the 80th percentile of the graph's degree distribution. The scaling term maps the node degrees to a range between 0 and 1 and caps it at the $80$-th percentile. The learnable variable $\beta$ controls the strength of this scaling factor. We have modeled $g$ (\ref{eq:spring_g}) based on observations showing that the number of correctly predicted edge signs adjacent to a node correlates with the node degree. All the learnable parameters from {SPR} are organized in the vector
\begin{equation}
    \theta_{spring} = \begin{bmatrix}
        \theta_f && \beta
    \end{bmatrix}^{\top}.
\end{equation}

\subsection{{SPR-NN}: Neural Network Based Forces}
\label{SPR-NN}

In our second approach, we model $f$ based on the structure of a multi-layer perception (MLP). We define a shallow MLP with a single hidden layer as
\begin{equation}
    \text{MLP}(\theta_\text{MLP}, \mathbf{z}_{ij}) = \mathbf{W}_1 \times \text{ReLU}(\mathbf{W}_{0} \times \mathbf{z}_{ij} + \mathbf{b}_0) + \mathbf{b}_1
\end{equation}
where $\mathbf{W}_0$, $\mathbf{W}_1$, $\mathbf{b}_0$ and $\mathbf{b}_1$ are weight matrices and respectively bias vectors of the MLP. $\mathbf{z}_{ij}$ denotes a static edge feature vector, and ReLU denotes a rectified linear unit activation function. Finally
\begin{equation}
    \theta_\text{MLP} = \begin{bmatrix}
        \mathbf{W}_{0} && \mathbf{W}_{1} && \mathbf{b}_0 && \mathbf{b}_1
    \end{bmatrix}^{\top}.
\end{equation}
The function $f$ is made of three equivalent but individually parameterized neural networks
\begin{equation}
    f( \theta_f, \mathbf{z}_{ij}) =
    \begin{cases}
         \text{MLP}(\theta_{0}, \mathbf{z}_{ij}) &
         \text{if $\sigma^{\prime}(i, j)$ is 0} \\
            \text{MLP}(\theta_{1}, \mathbf{z}_{ij}) &
         \text{if $\sigma^{\prime}(i, j)$ is 1} \\
          \text{MLP}(\theta_{2}, \mathbf{z}_{ij}) &
         \text{if $\sigma^{\prime}(i, j)$ is -1} \\
    \end{cases},
\label{eq:spring_scalarf_nn}
\end{equation}
where \begin{equation}
    \theta_f = \begin{bmatrix}
       \theta_{0} && \theta_{1} && \theta_{2}
    \end{bmatrix}^{\top}.
\end{equation}
The size of the hidden layer for each MLP is seven, hence $\mathbf{W}_0 \in \mathbb{R}^{7 \times 7}$, $\mathbf{W}_1 \in \mathbb{R}^{7}$, $\mathbf{b}_0 \in \mathbb{R}^{7}$, $\mathbf{b}_1 \in \mathbb{R}$. The edge specific input vector $\mathbf{z}_{ij}$ is defined as
\begin{equation}
    \mathbf{z}_{ij} = \begin{bmatrix}
        d(\mathbf{x}_i, \mathbf{x}_j ) &
        \text{deg}_i &
        \text{deg}_j &
        \text{neg}_i &
        \text{neg}_j &
        \text{pos}_i &
        \text{pos}_j
    \end{bmatrix}^{\top}.
\end{equation}
The variables $\text{neg}_i$ and $\text{pos}_i$ represent the fractions of negative and positive edges connected to node $i$. Given that some edges are unknown during training, $\text{neg}_i$ and $\text{pos}_i$ are crucial for our analysis. Naturally, the features are constructed without knowing the true sign distribution function $\sigma$.

The function $g$ is
\begin{equation}
    g(\theta_g, \mathbf{z}_i) = \text{MLP}(\theta_g, \mathbf{z}_i)
\end{equation}
with $\mathbf{W}_0 \in \mathbb{R}^{3 \times 3}$, $\mathbf{W}_1 \in \mathbb{R}^{3}$, $\mathbf{b}_0 \in \mathbb{R}^{3}$, $\mathbf{b}_1 \in \mathbb{R}$.
Hence the complete parameter vector
\begin{equation}
    \theta_{nn} = \begin{bmatrix}
       \theta_{g} && \theta_{f}
    \end{bmatrix}^{\top}
\end{equation}
contains 8 weight matrices and and 8 bias vectors, totalling in 184 parameters.

\section{Simulation and Training}
\label{Simulation}

The second-order ODE of the embedding vector $\mathbf{x}$ is equivalent to the equation of motion with a constant mass of one. Hence, acceleration is equal to force and
\begin{equation}
    \label{eq:second-2}
    \frac{d^2\mathbf{x}_i}{dt^2} = \frac{d\mathbf{v}_i}{dt} = \mathbf{f}_i,
\end{equation}
which can be solved by the Euler method. In this section, we describe the forward simulation and the optimization process, which leverages a differentiable formulation of the forward simulation to learn $\theta_\text{spring}$ and $\theta_{nn}$.

\subsection{Forward Simulation}

We introduce the time-dependent position, velocity and force tensors
\begin{equation}
    \mathbf{X}(t) = \begin{bmatrix}
        \mathbf{x}_0^{\top}(t) \\ \mathbf{x}_1^{\top}(t) \\ \dots \\ \mathbf{x}_N^{\top}(t)
    \end{bmatrix} \in  \mathbb{R}^{N \times k}, \quad
    \mathbf{V}(t) = \begin{bmatrix}
        \mathbf{v}_0^{\top}(t) \\ \mathbf{v}_1^{\top}(t) \\ \dots \\ \mathbf{v}_N^{\top}(t)
    \end{bmatrix} \in  \mathbb{R}^{N \times k}
\end{equation}
\begin{equation}
    \text{ and }  \quad
    \mathbf{F}(t) = \begin{bmatrix}
        \textbf{f}_0^{\top}(t) \\ \textbf{f}_1^{\top}(t) \\ \dots \\ \textbf{f}_N^{\top}(t)
    \end{bmatrix} \in  \mathbb{R}^{N \times k}.
\end{equation}
The functions $f$ and $g$ of the forces $\mathbf{f}$ can be either chosen as defined for {SPR} or {SPR-NN}. The equations to advance $\mathbf{V}(t)$ and $\mathbf{X}(t)$ in time are
\begin{equation}
    \mathbf{V}(t + dt) = (1 - d) \cdot \mathbf{V}(t) + dt \cdot \mathbf{F}(t)
\end{equation}
and
\begin{equation}
    \mathbf{X}(t + dt) = \mathbf{X}(t) + dt \cdot \mathbf{V}(t),
\end{equation}
where $d$ is a linear damping factor, which increases the numerical stability. Using the state vector notation $\mathbf{S}(t)=\left[\mathbf{X}(t) \mathbf{V}(t)\right]^{\top}$, we define the update function $\Phi$ which advances the state by a time step $dt$ as
\begin{equation}
    \Phi \bigl( \mathbf{S}(t) \bigr) = \begin{bmatrix}
        1 & 0 \\ 0 & (1-d)
    \end{bmatrix}\mathbf{S}(t) + dt \begin{bmatrix} \mathbf{V}(t) \\  \mathbf{F}(t)\end{bmatrix}.
\end{equation}
Hence a forward simulation is a composition of $\Phi$ functions \begin{equation}
\label{eq:fwd-simulation}
    \mathbf{S}(n ) = (\overbrace{\Phi \circ \dots \circ \Phi}^{n \text{-times}})(\mathbf{S}(0) )
\end{equation}
where $n$ denotes the number of simulation steps. The initial state of the system, $\mathbf{S}(0)$, is defined as described in equations \ref{eq:c3} and \ref{eq:c4}. The time complexity of the entire forward simulation is $O(n (N  k + M  k))$ as it is a chain of  GSN layers, each with a complexity of $O(N  k + M  k)$.

\begin{figure}[h]
\centering
\includegraphics[width=0.7\textwidth]{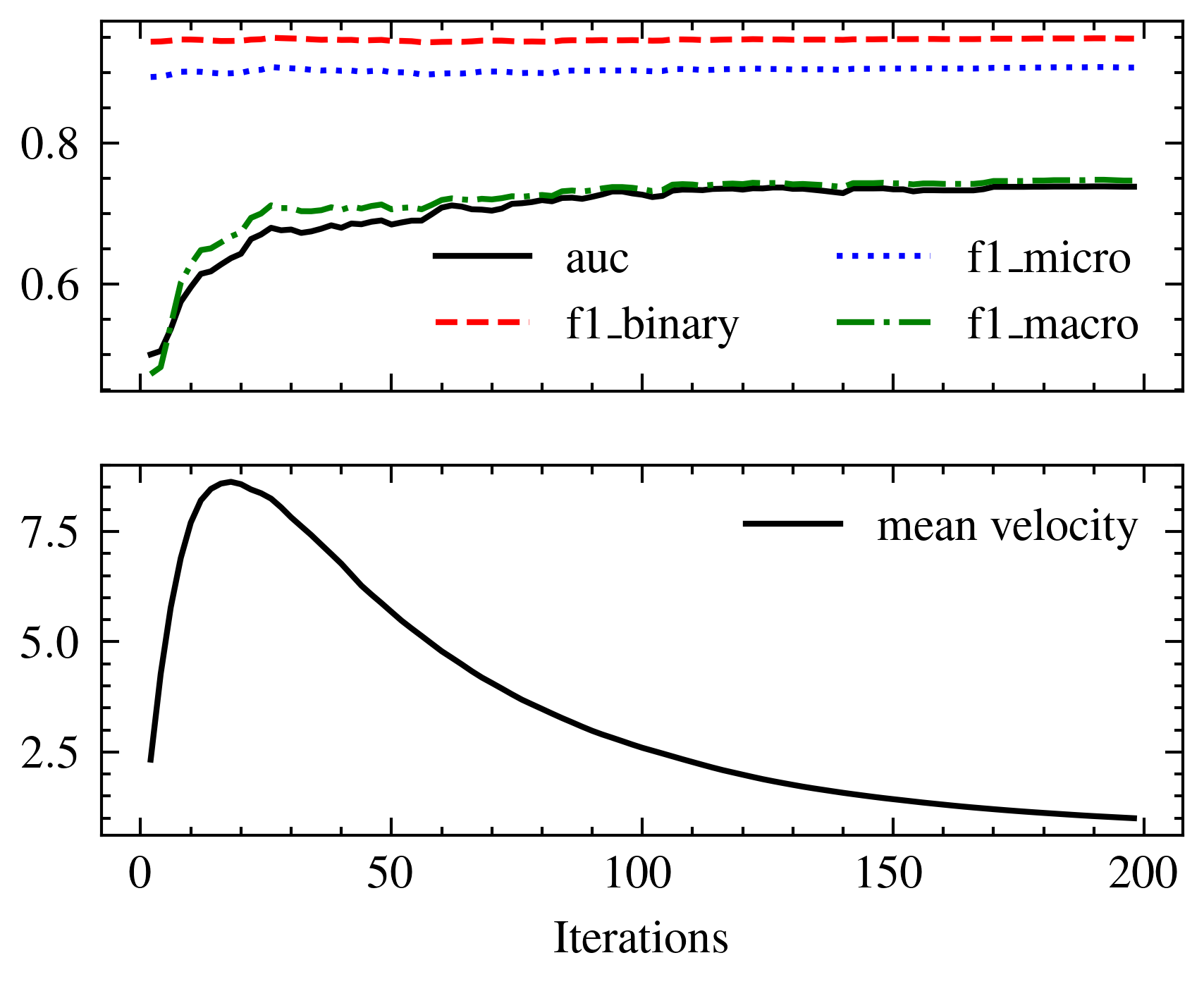}
\caption{Forward simulation of {SPR-NN} on BitcoinAlpha. The mean velocity is equivalent to $\frac{d\mathbf{x}}{dt}$.}
\label{fig:forward-nn}
\end{figure}

\subsection{Parameter Learning}

\begin{figure}[h]
\centering
\includegraphics[width=0.7\textwidth]{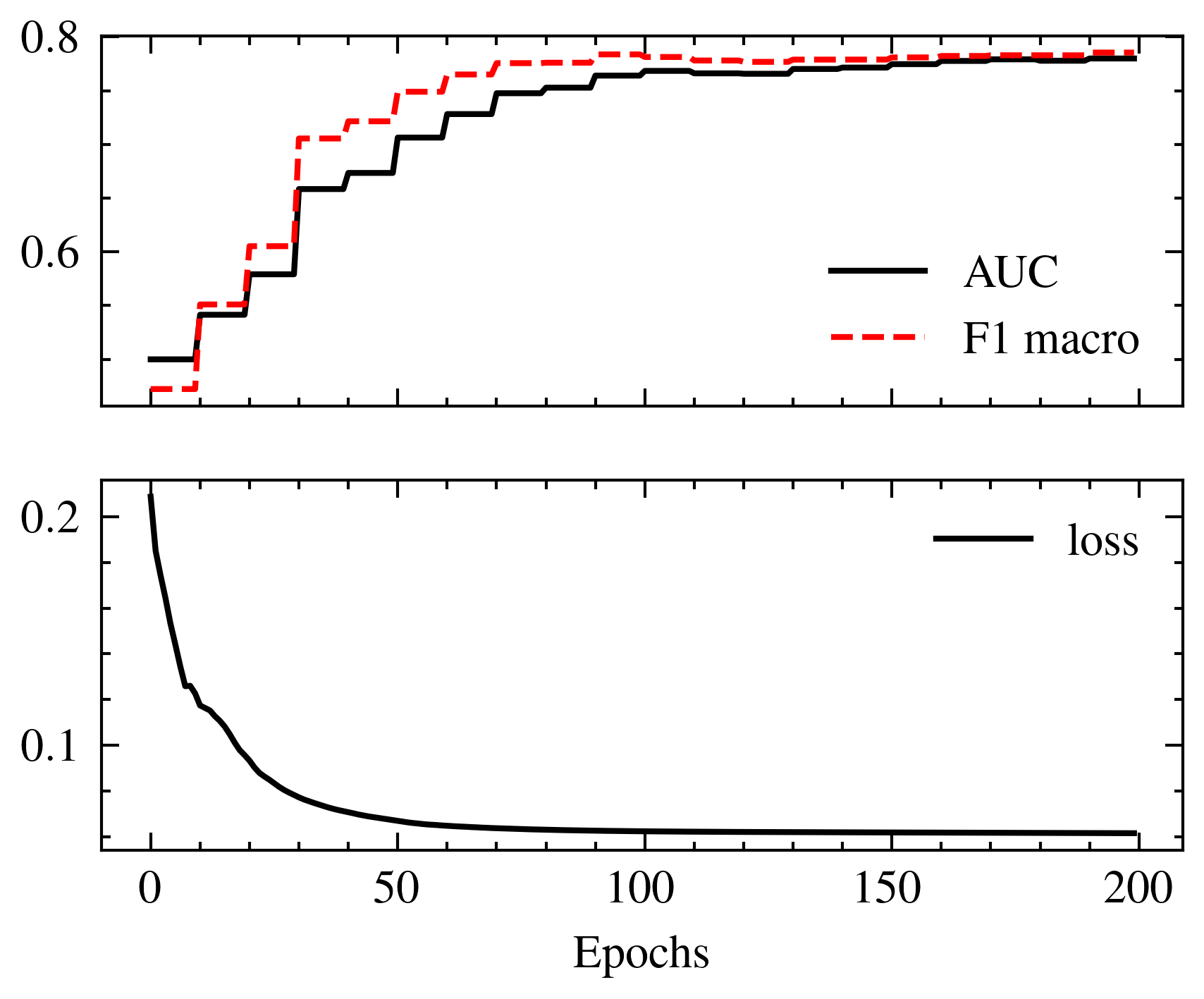}
\caption{Training of {SPR-NN} on BitcoinAlpha.}
\label{fig:train-nn-1}
\end{figure}

To learn parameters $\theta_\text{spring}$ and $\theta_\text{nn}$, we leverage a gradient based optimization method, hence a differentiable loss function needs to be found. Sign prediction is a discrete task, however we can map the edge lengths to the interval $[0, 1]$ using a logistic function with a threshold $\mu$
\begin{equation}
    \widehat{\sigma}(u, v) = \left(1 + e^{ \|{\mathbf{x}_u - \mathbf{x}_v}\|_2 - \mu}\right)^{-1}
\end{equation}
transforming it into a prediction task with a notion of confidence. This formulation allows us to define the loss as the weighted mean squared difference between the prediction $\widehat{\sigma}$ and the ground truth $\sigma$
\begin{equation}
    L(G, \textbf{S}(t)) = \sum_{(u, v) \in E} (\sigma(u, v) - \widehat{\sigma}(u, v))^2 \cdot \omega(u, v),
\end{equation}
with normalization function
\begin{equation}
\omega(u, v) = \begin{cases}
\frac{1}{|E^{+}|} & \text{if } \sigma^{\prime}(u, v) = 1 \\
\frac{1}{|E^{-}|} & \text{if } \sigma^{\prime}(u, v) = -1
\end{cases},
\end{equation}
which ensures an equal contribution of positive and negative predictions to the loss function, which is important in unbalanced datasets (see Tab. \ref{tab:data}). Finally, the gradient of the loss function with respect to the parameters $\theta$ we wish to learn is given by
\begin{equation}
    \nabla L  = \frac{\partial L}{\partial \mathbf{S}(n)} \frac{\partial \mathbf{S}(n)}{\partial \theta} + \frac{\partial L}{\partial \theta},
\end{equation}

where in our case $\theta$ could be either $\theta_\text{spring}$ or $\theta_\text{nn}$. With modern Automatic Differentiation (AD) libraries, $\nabla L$ can be obtained directly by calling a gradient computation function, with the loss function including the entire forward simulation as a parameter. In our case, we use the JAX library \citep{jax2018github}. The gradient needs to be back-propagated over the full chain of forward solver steps, leading to exploding and vanishing gradients \citep{thuerey2021physics}. To address the exploding gradients problem we clip the gradient to a range $[-1, 1]$ \citep{zhang2019gradient}. We then use the Adam optimization algorithm \citep{kingma2014adam} to improve the parameters gradually. The decrease in loss and increase in the classification metrics AUC and F1-Macro can be observed in Figures \ref{fig:forward-nn} and \ref{fig:train-nn-1}. A drawback of applying AD on an iterative solver to obtain $\nabla L$ is a memory load that increases linearly with the number of forward iteration steps.

\section{Experiments}
\label{results}

\subsection{Datasets}
We evaluate our methods on four empirical signed networks.
\begin{itemize}
    \item BitcoinAlpha and BitcoinOTC are networks where users can indicate their level of trust for other users in integer values between -10 and 10 \citep{kumar2016edge}. We process the levels by taking the sign of the level as the sign of the edge.
    \item Slashdot is a platform that features user-edited content, where users can mark other users as friends or foes \citep{leskovec2009community}.
    \item Epinions is a social network where users can indicate their trust and distrust against other users     \citep{richardson2003trust}.
\end{itemize}

\begin{table}[t]
\caption{Proportion of positive edges, number of edges and nodes of each network in the test datasets.}
\label{tab:data}
    \centering
    \begin{tabular}{l|l|l|l|l}
     & Positive Edges &
     Edges $M$ & Nodes $N$ \\
    \toprule
        \parbox{12mm}{Bitcoin \\Alpha} & 0.900  & 24186 & 3783\\
        \midrule
        \parbox{12mm}{Bitcoin \\OTC}  & 0.848  & 35592 & 5881 \\
        \midrule
        Slashdot & 0.773 & 516575 & 77357\\
        \midrule
        Epinions & 0.829 & 841372 & 131828\\
    \bottomrule
    \end{tabular}
\end{table}

The characteristics of each network are shown in Table. \ref{tab:data}.
We chose these datasets because they represent different types of signed networks, cover network sizes from small to large, and have been regularly used for comparisons by other authors.
We convert directed into undirected graphs. For each edge $(i, j)$ in the undirected graph, we add the reverse edge from $(j, i)$ with an identical sign. If the reverse edge already exists in the dataset and there is a disagreement in sign, we prioritize the negative connection, converting both to negative as it is common practice \citep{wang2017signed, derr2018signed}.

\subsection{Compared Methods}

We compare SPR and SPR-NN with state-of-the-art methods for link sign prediction.

\begin{itemize}
  \item SIGNET introduces a random walk method to preserve structural equilibrium in signed networks \citep{islam2018signet}.
  \item SGCN extends the Graph Convolutional Network (GCN) framework to accommodate signed networks, drawing on balance theory principles \citep{derr2018signed}.
  \item SiGAT integrates graph motifs specific to signed networks into the Graph Attention Network (GAT) model, leveraging balance theory and status theory \citep{huang2019signed}.
  \item SNEA employs an attention mechanism rooted in balance theory to evaluate the importance of various neighbors during information dissemination \citep{li2020learning}.
  \item SGDNET utilizes a random walk technique to diffuse information across the network\citep{jung2020signed}.
  \item SDGNN offers a comprehensive approach to reconstruct the signs of links simultaneously, the directions of links, and signed directed triangles, enhancing the modeling of signed networks \citep{huang2021sdgnn}.
  \item SLGNN leverages spectral graph theory and graph signal processing to extract and combine low- and high-frequency information from positive and negative links. It uses a self-gating mechanism to dynamically adjust the influence of this information \citep{li2023signed}.
\end{itemize}

For training SPR-NN and SPR, we use the following settings: embedding dimension $k = 64$, damping factor $d = 0.05$, temporal resolution $dt = 0.005$ and threshold $\mu=2.5$, with a learning rate of 0.03, which we train over 200 epochs with 120 Euler integration iterations each. We have not used any hyperparameter tuning to improve the parameters systematically.

\subsection{Classification Performance}

% \begin{sidewaystable}
\begin{table*}[h]{\setlength{\tabcolsep}{1mm}}
\fontsize{7}{7}\selectfont
 \centering
    \caption{Link sign prediction results (mean±std). Results for all methods except SPR and SPR-NN are reproduced from \citep{li2023signed}. The best are bolded and second best underlined.}
  \begin{tabular}{p{0.2cm}|p{1.0cm}|c|c|c|c|c|c|c|c|c}
    \hline
    &Metrics & SIGNET & SGCN & SiGAT & SNEA & SGDNET & SDGNN & SLGNN & SPR & SPR-NN \\
    \hline
    \multirow{5}{*}{\rotatebox[origin=c]{90}{Bit. Alpha }} & F1-MI & 92.82{±\tiny0.31} & 92.65{±\tiny0.30} & 92.08{±\tiny0.24} & \underline{92.99}{±\tiny0.29} & 92.30{±\tiny0.22} & 92.55{±\tiny0.32} & \textbf{94.28}{±\tiny0.51} & 91.62 {±\tiny0.33} & 90.99 {±\tiny0.33}
    \\
    & F1-MA & 74.91{±\tiny1.33} & 75.67{±\tiny1.46} & 72.93{±\tiny0.92} & 76.46{±\tiny1.19} & 75.22{±\tiny0.92} & \underline{78.04}{±\tiny1.03} & \textbf{83.61}{±\tiny0.94} & 73.0 {±\tiny1.39} & 76.10 {±\tiny0.95}
    \\
    & F1-WT & 91.92{±\tiny0.40} & 91.98{±\tiny0.41} & 91.20{±\tiny0.27} & 92.29{±\tiny0.36} & 91.72{±\tiny0.25} & \underline{92.36}{±\tiny0.34} & \textbf{94.22}{±\tiny0.42} & 91.05{±\tiny0.40} & 91.24{±\tiny0.39}
    \\
    & F1-BI & 96.11{±\tiny0.17} & 95.99{±\tiny0.16} & 95.70{±\tiny0.13} & 96.19{±\tiny0.16} & 95.79{±\tiny0.12} & \underline{95.89}{±\tiny0.17} & \textbf{96.83}{±\tiny0.29} & 95.42{±\tiny0.18} & 95.07{±\tiny0.19}
    \\
    & AUC-P & 92.02{±\tiny0.56} & 92.31{±\tiny0.82} & 87.71{±\tiny0.78} & \underline{92.70}{±\tiny0.58} & 89.41{±\tiny0.34} & 91.88{±\tiny0.52} & \textbf{95.08}{±\tiny0.34} & 85.03{±\tiny1.15}  &89.35{±\tiny0.10}
    \\
    & AUC-L & 70.29{±\tiny1.26} & 71.99{±\tiny1.74} & 69.05{±\tiny0.97} & 72.31{±\tiny1.20} & 72.14{±\tiny1.25} & 76.75{±\tiny1.21} & \textbf{82.88}{±\tiny0.82} & 70.4{±\tiny1.54} & \underline{76.67}{±\tiny0.14}
    \\
    \hline
    \multirow{5}{*}{\rotatebox[origin=c]{90}{Bit. OTC }} & F1-MI & 90.68{±\tiny0.29} & 91.72{±\tiny0.25} & 90.54{±\tiny0.26} & \underline{92.28}{±\tiny0.27} & 91.80{±\tiny0.47} & 92.26{±\tiny0.28} & \textbf{94.48}{±\tiny0.33} & 90.85{±\tiny0.23} & 91.97{±\tiny0.43}
    \\
    & F1-MA & 79.62{±\tiny0.71} & 82.37{±\tiny0.57} & 79.11{±\tiny0.87} & 83.45{±\tiny0.66} & 83.54{±\tiny0.90} & \underline{84.57}{±\tiny0.65} & \textbf{88.99}{±\tiny0.56} & 80.86{±\tiny0.75} & 84.12{±\tiny0.87}
    \\
    & F1-WT & 90.09{±\tiny0.33} & 91.32{±\tiny0.27} & 89.88{±\tiny0.35} & 91.87{±\tiny0.30} & 91.67{±\tiny0.46} & \underline{92.16}{±\tiny0.31} & \textbf{94.41}{±\tiny0.31} & 90.46{±\tiny0.23} & 91.95{±\tiny0.41}
    \\
    & F1-BI & 94.63{±\tiny0.16} & 95.21{±\tiny0.14} & 94.56{±\tiny0.13} & \underline{95.54}{±\tiny0.15} & 95.20{±\tiny0.28} & 95.47{±\tiny0.16} & \textbf{96.77}{±\tiny0.20} & 94.69{±\tiny0.12} & 95.28{±\tiny0.25}
    \\
    & AUC-P & 90.78{±\tiny0.73} & 92.31{±\tiny0.28} & \underline{94.70}{±\tiny0.58} & 93.84{±\tiny0.52} & 92.97{±\tiny0.67} & 94.19{±\tiny0.26} & \textbf{96.87}{±\tiny0.39} & 89.18{±\tiny0.54} & 92.61{±\tiny0.56}
    \\
    & AUC-L & 76.50{±\tiny0.77} & 79.67{±\tiny1.63} & 75.85{±\tiny1.12} & 80.14{±\tiny1.20} & 82.49{±\tiny0.87} & 83.73{±\tiny0.89} & \textbf{87.95}{±\tiny0.27} & 78.54{±\tiny0.71s} & \underline{84.0}{±\tiny0.73}\\
    \hline
    \multirow{5}{*}{\rotatebox[origin=c]{90}{Slashdot }} & F1-MI & 83.28{±\tiny0.10} & 83.17{±\tiny0.17} & 82.93{±\tiny0.08} & 84.74{±\tiny1.00} & 84.22{±\tiny0.12} & 84.05{±\tiny1.08} & \textbf{87.83}{±\tiny0.09} & 83.80{±\tiny0.09} & \underline{85.33}{±\tiny0.16}
    \\
    & F1-MA & 76.22{±\tiny0.15} & 75.73{±\tiny0.38} & 75.73{±\tiny0.11} & 78.45{±\tiny0.12} & 77.31{±\tiny0.32} & 77.98{±\tiny0.26} & \textbf{83.27}{±\tiny0.13} & 72.85{±\tiny0.25} & \underline{78.88}{±\tiny0.33}
    \\
    & F1-WT & 82.65{±\tiny0.11} & 82.40{±\tiny0.23} & 82.29{±\tiny0.08} & 84.23{±\tiny0.09} & 83.53{±\tiny0.18} & 83.72{±\tiny0.19} & \textbf{87.61}{±\tiny0.10} & 82.31{±\tiny0.15} & \underline{85.26}{±\tiny0.19}
    \\
    & F1-BI & 89.18{±\tiny0.07} & 89.17{±\tiny0.10} & 88.95{±\tiny0.06} & 90.09{±\tiny0.07} & 89.82{±\tiny0.06} & 89.54{±\tiny0.12} & \textbf{92.01}{±\tiny0.06} & \underline{90.90}{±\tiny0.05} & 90.55{±\tiny0.09}
    \\
    & AUC-P & 84.42{±\tiny0.08} & 87.75{±\tiny0.30} & 89.35{±\tiny0.10} & 90.00{±\tiny0.10} & \underline{88.98}{±\tiny0.15} & 88.59{±\tiny0.11} & \textbf{93.22}{±\tiny0.05} & 86.80{±\tiny0.08} & 88.90{±\tiny0.26}
    \\
    & AUC-L & 74.62{±\tiny1.13} & 73.89{±\tiny0.44} & 74.17{±\tiny0.12} & 76.87{±\tiny0.11} & 75.43{±\tiny0.46} & 76.95{±\tiny0.27} & \textbf{82.15}{±\tiny0.16} & 70.30{±\tiny0.05} & \underline{78.63}{±\tiny0.45}\\
    \hline
    \multirow{5}{*}{\rotatebox[origin=c]{90}{Epinions}} & F1-MI & 91.25{±\tiny0.05} & 92.34{±\tiny0.05} & 90.75{±\tiny0.06} & 92.48{±\tiny0.05} & 91.76{±\tiny0.44} & \underline{92.71}{±\tiny0.40} & \textbf{94.40}{±\tiny0.08} & 90.49{±\tiny0.12} & 91.33{±\tiny0.13}
    \\
    & F1-MA & 82.29{±\tiny0.09} & 85.12{±\tiny0.05} & 81.74{±\tiny0.17} & 85.42{±\tiny0.08} & 83.93{±\tiny0.48} & \underline{86.02}{±\tiny0.09} & \textbf{89.44}{±\tiny0.18} & 82.04{±\tiny0.08} & 83.38{±\tiny0.65}
    \\
    & F1-WT & 90.71{±\tiny0.05} & 92.04{±\tiny0.05} & 90.31{±\tiny0.06} & 92.20{±\tiny0.05} & 91.43{±\tiny0.06} & \underline{92.48}{±\tiny0.05} & \textbf{94.28}{±\tiny0.09} & 90.16{±\tiny0.08} & 90.79{±\tiny0.23}
    \\
    & F1-BI & 94.89{±\tiny0.03} & 95.48{±\tiny0.07} & 94.56{±\tiny0.04} & 95.56{±\tiny0.03} & 95.14{±\tiny0.03} & \underline{95.69}{±\tiny0.02} & \textbf{96.68}{±\tiny0.04} & 94.36{±\tiny0.05} & 94.88{±\tiny0.08}
    \\
    & AUC-P & 93.19{±\tiny0.06} & 95.17{±\tiny0.03} & 94.35{±\tiny0.05} & 95.30{±\tiny0.07} & 94.42{±\tiny0.06} & \underline{95.06}{±\tiny0.06} & \textbf{97.02}{±\tiny0.06} & 93.38{±\tiny0.08} & 92.68{±\tiny1.15}
    \\
    & AUC-L & 78.99{±\tiny0.08} & 82.75{±\tiny0.09} & 79.16{±\tiny0.28} & 83.08{±\tiny0.09} & 81.54{±\tiny0.32} & \underline{84.00}{±\tiny0.15} & \textbf{87.79}{±\tiny0.36} & 80.06{±\tiny0.17} &  80.97{±\tiny1.33} \\
    \hline
  \end{tabular}

    \label{tab:scores}
\end{table*}
% \end{sidewaystable}

We benchmark the accuracy of our methods against other methods based on Micro-F1, Macro-F1, Weighted-F1, Binary-F1, AUC-P and AUC-L scores. AUC-L is the AUC score evaluated on the binary predictions from the logistic classifier, whereas AUC-P is evaluated on the prediction probabilities.
Contrary to AUC and Macro-F1, Micro-F1 and Binary-F1 do not consider class imbalance.
As the datasets are all strongly unbalanced, as seen in Table \ref{tab:data}, good scores for Micro and Binary-F1 can be achieved with classifiers that would always predict a positive sign.

The results are reported in Table \ref{tab:scores}.
We reproduce the results of other methods from  \citep{li2023signed}. All methods, including ours, use an embedding dimension of $k=64$ and a training split of 80\%.
The parameters used for the test results on all datasets except BitcoinAlpha were trained on BitcoinAlpha. For the test on BitcoinAlpha, we have trained our models on BitcoinOTC, showing the generalizability of our method.
This highlights how training can be done on a dataset different from the one the embedding generation is computed on.
Although, SLGNN consistently obtains the best result, our methods {SPR} and {SPR-NN} fit into the scoreboard of the compared methods. In particular, {SPR-NN} often ranks second and obtains results very close to SDGNN.

\subsection{Runtime}

\begin{figure}[h]
\centering
\includegraphics[width=0.8\textwidth]{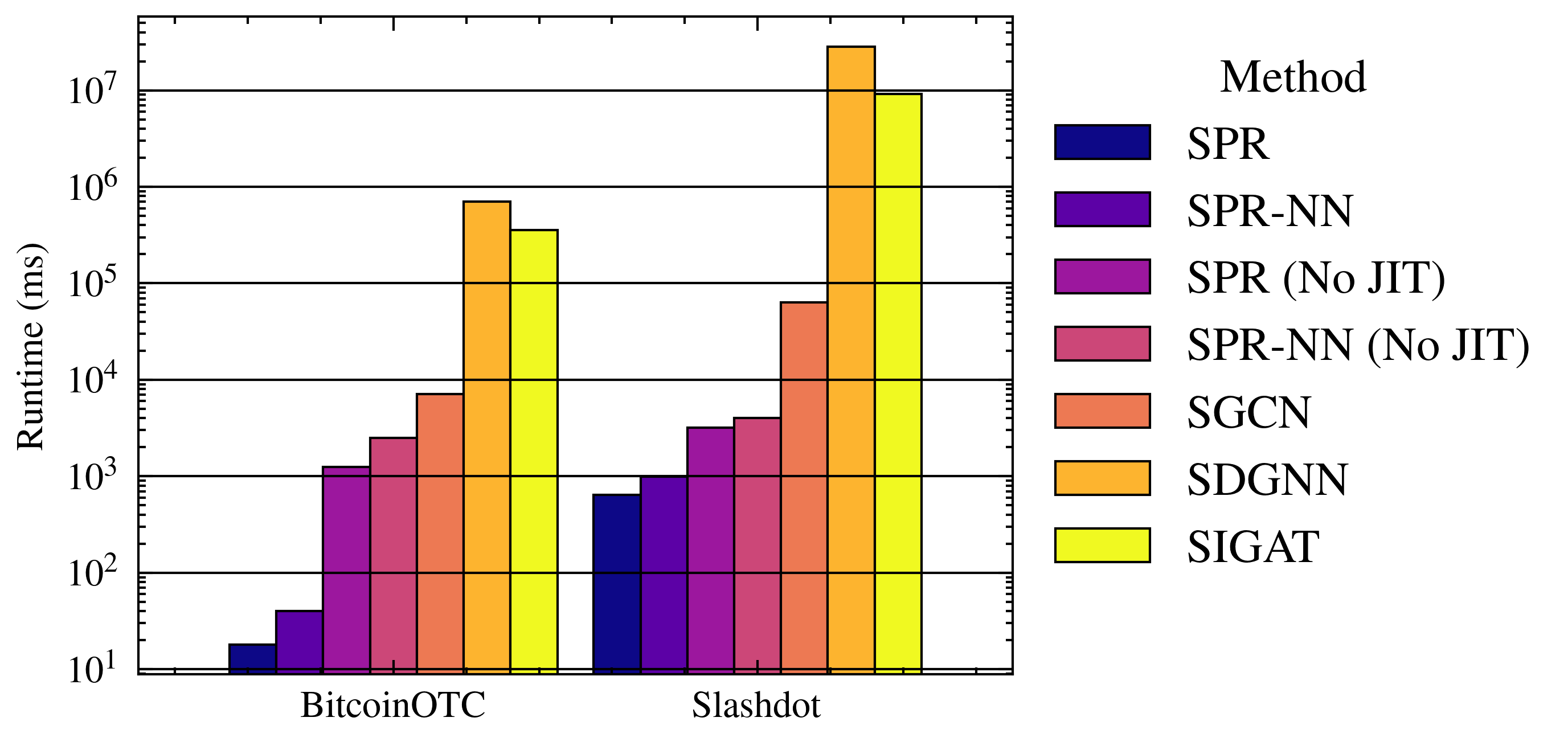}
\caption{Runtime comparisons between different methods and datasets. Measurements are an average of over seven executions. Lower is better.}
\label{fig:performance}
\end{figure}

The embedding generation time of {SPR} and {SPR-NN} are measured and compared with {SGCN} \citep{derr2018signed}, {SiGAT} \citep{huang2019signed} and {SDGNN} \citep{huang2021sdgnn}, where we rely on default settings of the reference implementations. We were not able to access the code for SLGNN. All methods are run on a system with the following specifications: AMD EPYC 7702 64-Core Processor: 16GB CPU RAM, NVIDIA Tesla T4: 80Gb GPU RAM, Ubuntu 20.04.
Our methods are implemented using Python version 3.12.4, JAX version 0.4.31 and the Adam optimization method is from Optax 0.2.3.

All compared implementations {SGCN}, {SiGAT}, and {SDGNN} use Pytorch, while our methods use JAX. JAX and Pytorch are fully GPU-accelerated array computation libraries with just-in-time compilation (JIT) features, providing near-optimal performance \citep{jax2018github, ansel2024pytorch}. We measure only the node embedding generation time; graph processing computation of node and edge level features are excluded. In the case of {SPR} and {SPR-NN} node embedding generation time is equivalent to the duration it takes to solve the ODEs. For other methods, this means learning the parameters of a Graph Neural Network capable of generating the embeddings for a specific dataset. We measure the execution times in milliseconds on BitcoinOTC and Slashdot, representing a medium and large dataset. The graph sizes can be found in Table \ref{tab:data} and the results are shown in Figure \ref{fig:performance}.
Due to the potential differing performance in JIT capabilities of PyTorch and JAX, we include measurements where the JIT feature of JAX is disabled. In the case of BitcoinOTC, JIT drastically increases performance, an effect which is less prevalent in the case of Slashdot. JIT likely removes additional overhead, which is dominant in the case of the relatively small graph BitcoinAlpha but not so on Slashdot.
The JIT compiled {SPR-NN} has a speedup factor of 63 over {SGCN} on the Slashdot dataset. Compared to the implementation of SDGNN, the speedup is 28654.
Compared to SGCN, which has a similar accuracy performance to SPR and SPR-NN, the runtime performance improvements likely stem from the following sources:
\begin{itemize}
    \item Better performance of the GSN layer compared to the GCN layer. This effect, however, is mostly hidden by GPU parallelization, as the layer operations can be executed in parallel.
    \item {SPR} and {SPR-NN} use only a single layer, whereas SGCN uses 32 consecutive layers. As the computation of the layers is serial, it cannot be parallelized.
    \item SGCN trains over 100 epochs, similarly to our methods, which use 120 Euler integration steps. The training process of SGCN, however, relies on backpropagation, which doubles its runtime compared to ours \citep{blakely2021time}.
\end{itemize}

\section{Conclusion}

In this paper, we investigate node feature vector generation for signed networks to optimize link sign prediction with a logistic classifier based on node vector distances. We propose an efficient message passing layer architecture called Graph Spring Network (GSN), which we combine with the Graph Neural ODE formalism. Based on this framework, we then propose the {SPR} method, which closely resembles the physical model of a spring, where each edge represents a spring that can store or release energy, generating node displacement. We extend the idea by replacing the spring mechanics with small artificial neural networks in {SPR-NN}. We train the specific dynamics of the ODEs for {SPR} and {SPR-NN} on a single dataset, allowing us to generate the node embeddings on novel datasets only by solving the ODEs using the Euler methods.
We then perform experiments on four real-world signed networks and show that our proposed method {SPR-NN} performs similarly to other state-of-the-art feature vector generation methods in terms of accuracy. Furthermore, our method is significantly faster in embedding generation than the compared methods.
Compared to other methods, our method marks the edges one wants to infer as neutral during embedding generation.
When the sign of a new set of edges is asked, the embedding has to be regenerated, which is not necessarily the case in other methods.
However, as our method is several orders of magnitude faster, regenerating the embeddings is not a problem. In further work {SPR} and {SPR-NN} could be adapted for directed networks and {SPR-NN} could be trained on more node features. Finally, the Graph Spring Network Layer (GSN) could be used for other node embedding generation tasks.

\bibliographystyle{unsrtnat}
\bibliography{sn-bibliography}  %%% Uncomment this line and comment out the ``thebibliography'' section below to use the external .bib file (using bibtex) .

\begin{thebibliography}{49}
\providecommand{\natexlab}[1]{#1}
\providecommand{\url}[1]{\texttt{#1}}
\expandafter\ifx\csname urlstyle\endcsname\relax
  \providecommand{\doi}[1]{doi: #1}\else
  \providecommand{\doi}{doi: \begingroup \urlstyle{rm}\Url}\fi

\bibitem[Leskovec et~al.(2010)Leskovec, Huttenlocher, and
  Kleinberg]{leskovec2010signed}
Jure Leskovec, Daniel Huttenlocher, and Jon Kleinberg.
\newblock Signed networks in social media.
\newblock In \emph{{Proceedings of the SIGCHI Conference on Human Factors in
  Computing Systems}}, pages 1361--1370, 2010.

\bibitem[Huang et~al.(2023)Huang, Xie, Cao, Shen, Zhang, Xia, and
  Cheng]{huang2023negative}
Junjie Huang, Ruobing Xie, Qi~Cao, Huawei Shen, Shaoliang Zhang, Feng Xia, and
  Xueqi Cheng.
\newblock Negative can be positive: Signed graph neural networks for
  recommendation.
\newblock \emph{{Information Processing \& Management}}, 60\penalty0
  (4):\penalty0 103403, 2023.

\bibitem[Cartwright and Harary(1956)]{cartwright1956structural}
Dorwin Cartwright and Frank Harary.
\newblock Structural balance: a generalization of heider's theory.
\newblock \emph{{Psychological Review}}, 63\penalty0 (5):\penalty0 277, 1956.

\bibitem[Huang et~al.(2021)Huang, Shen, Hou, and Cheng]{huang2021sdgnn}
Junjie Huang, Huawei Shen, Liang Hou, and Xueqi Cheng.
\newblock Sdgnn: Learning node representation for signed directed networks.
\newblock In \emph{{Proceedings of the AAAI Conference on Artificial
  Intelligence}}, volume~35, pages 196--203, 2021.

\bibitem[Derr et~al.(2018)Derr, Ma, and Tang]{derr2018signed}
Tyler Derr, Yao Ma, and Jiliang Tang.
\newblock Signed graph convolutional networks.
\newblock In \emph{{2018 IEEE International Conference on Data Mining (ICDM)}},
  pages 929--934. IEEE, 2018.

\bibitem[Poli et~al.(2019)Poli, Massaroli, Park, Yamashita, Asama, and
  Park]{poli2019graph}
Michael Poli, Stefano Massaroli, Junyoung Park, Atsushi Yamashita, Hajime
  Asama, and Jinkyoo Park.
\newblock Graph neural ordinary differential equations, 2019.
\newblock Preprint at \url{https://arxiv.org/abs/1911.07532}.

\bibitem[Rusch et~al.(2022)Rusch, Chamberlain, Rowbottom, Mishra, and
  Bronstein]{rusch2022graph}
T~Konstantin Rusch, Ben Chamberlain, James Rowbottom, Siddhartha Mishra, and
  Michael Bronstein.
\newblock Graph-coupled oscillator networks.
\newblock In \emph{{International Conference on Machine Learning}}, pages
  18888--18909. PMLR, 2022.

\bibitem[Sun et~al.(2024)Sun, Tian, Xiong, Zhang, Xiang, Jia, and
  Wang]{sun2024dynamise}
Haiting Sun, Peng Tian, Yun Xiong, Yao Zhang, Yali Xiang, Xing Jia, and Haofen
  Wang.
\newblock Dynamise: dynamic signed network embedding for link prediction.
\newblock \emph{{Machine Learning}}, 113\penalty0 (7):\penalty0 4037--4053,
  2024.

\bibitem[Liu(2022)]{liu2022lightsgcn}
Haoxin Liu.
\newblock Lightsgcn: Powering signed graph convolution network for link sign
  prediction with simplified architecture design.
\newblock In \emph{{Proceedings of the 45th International ACM SIGIR Conference
  on Research and Development in Information Retrieval}}, pages 2680--2685,
  2022.

\bibitem[Li et~al.(2020)Li, Tian, Zhang, and Chang]{li2020learning}
Yu~Li, Yuan Tian, Jiawei Zhang, and Yi~Chang.
\newblock Learning signed network embedding via graph attention.
\newblock In \emph{{Proceedings of the AAAI conference on Artificial
  Intelligence}}, volume~34, pages 4772--4779, 2020.

\bibitem[Islam et~al.(2018)Islam, Aditya~Prakash, and
  Ramakrishnan]{islam2018signet}
Mohammad~Raihanul Islam, B~Aditya~Prakash, and Naren Ramakrishnan.
\newblock Signet: Scalable embeddings for signed networks.
\newblock In \emph{{Advances in Knowledge Discovery and Data Mining: 22nd
  Pacific-Asia Conference, PAKDD 2018, Melbourne, VIC, Australia, June 3-6,
  2018, Proceedings, Part II 22}}, pages 157--169. Springer, 2018.

\bibitem[Ren et~al.(2024)Ren, Ding, Xu, and Zhang]{ren2024link}
Guojing Ren, Xiao Ding, Xiao-Ke Xu, and Hai-Feng Zhang.
\newblock Link prediction in multilayer networks via cross-network embedding.
\newblock In \emph{{Proceedings of the AAAI Conference on Artificial
  Intelligence}}, volume~38, pages 8939--8947, 2024.

\bibitem[Chung and Whang(2023)]{chung2023learning}
Chanyoung Chung and Joyce~Jiyoung Whang.
\newblock Learning representations of bi-level knowledge graphs for reasoning
  beyond link prediction.
\newblock In \emph{{Proceedings of the AAAI Conference on Artificial
  Intelligence}}, volume~37, pages 4208--4216, 2023.

\bibitem[Guha et~al.(2004)Guha, Kumar, Raghavan, and
  Tomkins]{guha2004propagation}
Ramanthan Guha, Ravi Kumar, Prabhakar Raghavan, and Andrew Tomkins.
\newblock Propagation of trust and distrust.
\newblock In \emph{{Proceedings of the 13th international conference on World
  Wide Web}}, pages 403--412, 2004.

\bibitem[Song and Meyer(2015)]{song2015link}
Dongjin Song and David~A Meyer.
\newblock Link sign prediction and ranking in signed directed social networks.
\newblock \emph{{Social Network Analysis and Mining}}, 5:\penalty0 1--14, 2015.

\bibitem[Leskovec et~al.(2009)Leskovec, Lang, Dasgupta, and
  Mahoney]{leskovec2009community}
Jure Leskovec, Kevin~J Lang, Anirban Dasgupta, and Michael~W Mahoney.
\newblock Community structure in large networks: Natural cluster sizes and the
  absence of large well-defined clusters.
\newblock \emph{{Internet Mathematics}}, 6\penalty0 (1):\penalty0 29--123,
  2009.

\bibitem[Perozzi et~al.(2014)Perozzi, Al-Rfou, and Skiena]{perozzi2014deepwalk}
Bryan Perozzi, Rami Al-Rfou, and Steven Skiena.
\newblock Deepwalk: Online learning of social representations.
\newblock In \emph{{Proceedings of the 20th ACM SIGKDD International Conference
  on Knowledge Discovery and Data Mining}}, pages 701--710, 2014.

\bibitem[Grover and Leskovec(2016)]{grover2016node2vec}
Aditya Grover and Jure Leskovec.
\newblock node2vec: Scalable feature learning for networks.
\newblock In \emph{{Proceedings of the 22nd ACM SIGKDD International Conference
  on Knowledge Discovery and Data Mining}}, pages 855--864, 2016.

\bibitem[Shi et~al.(2024)Shi, Hu, Zhao, He, Zhang, and Zhou]{shi2024structural}
Lei Shi, Bin Hu, Deng Zhao, Jianshan He, Zhiqiang Zhang, and Jun Zhou.
\newblock Structural information enhanced graph representation for link
  prediction.
\newblock In \emph{{Proceedings of the AAAI Conference on Artificial
  Intelligence}}, volume~38, pages 14964--14972, 2024.

\bibitem[Wang et~al.(2017)Wang, Tang, Aggarwal, Chang, and Liu]{wang2017signed}
Suhang Wang, Jiliang Tang, Charu Aggarwal, Yi~Chang, and Huan Liu.
\newblock Signed network embedding in social media.
\newblock In \emph{{Proceedings of the 2017 SIAM International Conference on
  Data Mining}}, pages 327--335. SIAM, 2017.

\bibitem[Chen et~al.(2018{\natexlab{a}})Chen, Qian, Liu, and
  Sun]{chen2018bridge}
Yiqi Chen, Tieyun Qian, Huan Liu, and Ke~Sun.
\newblock " bridge" enhanced signed directed network embedding.
\newblock In \emph{{Proceedings of the 27th ACM International Conference on
  Information and Knowledge Management}}, pages 773--782, 2018{\natexlab{a}}.

\bibitem[Fang et~al.(2023)Fang, Tan, and Wang]{fang2023signed}
Zhihong Fang, Shaolin Tan, and Yaonan Wang.
\newblock A signed subgraph encoding approach via linear optimization for link
  sign prediction.
\newblock \emph{{IEEE Transactions on Neural Networks and Learning Systems}},
  2023.

\bibitem[Veli{\v{c}}kovi{\'c} et~al.(2017)Veli{\v{c}}kovi{\'c}, Cucurull,
  Casanova, Romero, Lio, and Bengio]{velivckovic2017graph}
Petar Veli{\v{c}}kovi{\'c}, Guillem Cucurull, Arantxa Casanova, Adriana Romero,
  Pietro Lio, and Yoshua Bengio.
\newblock Graph attention networks, 2017.
\newblock Preprint at \url{https://arxiv.org/abs/1710.10903}.

\bibitem[Huang et~al.(2019)Huang, Shen, Hou, and Cheng]{huang2019signed}
Junjie Huang, Huawei Shen, Liang Hou, and Xueqi Cheng.
\newblock Signed graph attention networks.
\newblock In \emph{{Artificial Neural Networks and Machine Learning--ICANN
  2019: Workshop and Special Sessions: 28th International Conference on
  Artificial Neural Networks, Munich, Germany, September 17--19, 2019,
  Proceedings 28}}, pages 566--577. Springer, 2019.

\bibitem[Li et~al.(2023)Li, Qu, Tang, and Chang]{li2023signed}
Yu~Li, Meng Qu, Jian Tang, and Yi~Chang.
\newblock Signed laplacian graph neural networks.
\newblock In \emph{{Proceedings of the AAAI conference on Artificial
  Intelligence}}, volume~37, pages 4444--4452, 2023.

\bibitem[Lizurej et~al.(2023)Lizurej, Michalak, and
  Dziembowski]{lizurej2023manipulating}
Tomasz Lizurej, Tomasz Michalak, and Stefan Dziembowski.
\newblock On manipulating weight predictions in signed weighted networks.
\newblock In \emph{{Proceedings of the AAAI Conference on Artificial
  Intelligence}}, volume~37, pages 5222--5229, 2023.

\bibitem[Lee et~al.(2024)Lee, Lee, and Shin]{lee2024spear}
Dongjin Lee, Juho Lee, and Kijung Shin.
\newblock Spear and shield: Adversarial attacks and defense methods for
  model-based link prediction on continuous-time dynamic graphs.
\newblock In \emph{{Proceedings of the AAAI Conference on Artificial
  Intelligence}}, volume~38, pages 13374--13382, 2024.

\bibitem[Xhonneux et~al.(2020)Xhonneux, Qu, and Tang]{xhonneux2020continuous}
Louis-Pascal Xhonneux, Meng Qu, and Jian Tang.
\newblock Continuous graph neural networks.
\newblock In \emph{{International Conference on Machine Learning}}, pages
  10432--10441. PMLR, 2020.

\bibitem[Chen et~al.(2018{\natexlab{b}})Chen, Rubanova, Bettencourt, and
  Duvenaud]{chen2018neural}
Ricky~TQ Chen, Yulia Rubanova, Jesse Bettencourt, and David~K Duvenaud.
\newblock Neural ordinary differential equations.
\newblock In \emph{{Advances in Neural Information Processing Systems}},
  volume~31, 2018{\natexlab{b}}.

\bibitem[Huang et~al.(2020)Huang, Sun, and Wang]{huang2020learning}
Zijie Huang, Yizhou Sun, and Wei Wang.
\newblock Learning continuous system dynamics from irregularly-sampled partial
  observations.
\newblock In \emph{{Advances in Neural Information Processing Systems}},
  volume~33, pages 16177--16187, 2020.

\bibitem[Sanchez-Gonzalez et~al.(2019)Sanchez-Gonzalez, Bapst, Cranmer, and
  Battaglia]{sanchez2019hamiltonian}
Alvaro Sanchez-Gonzalez, Victor Bapst, Kyle Cranmer, and Peter Battaglia.
\newblock Hamiltonian graph networks with ode integrators, 2019.
\newblock Preprint at \url{https://arxiv.org/abs/1909.12790}.

\bibitem[Kang et~al.(2023)Kang, Zhao, Song, Wang, and Tay]{kang2023node}
Qiyu Kang, Kai Zhao, Yang Song, Sijie Wang, and Wee~Peng Tay.
\newblock Node embedding from neural hamiltonian orbits in graph neural
  networks.
\newblock In \emph{{International Conference on Machine Learning}}, pages
  15786--15808. PMLR, 2023.

\bibitem[Chamberlain et~al.(2021)Chamberlain, Rowbottom, Gorinova, Bronstein,
  Webb, and Rossi]{chamberlain2021grand}
Ben Chamberlain, James Rowbottom, Maria~I Gorinova, Michael Bronstein, Stefan
  Webb, and Emanuele Rossi.
\newblock Grand: Graph neural diffusion.
\newblock In \emph{{International Conference on Machine Learning}}, pages
  1407--1418. PMLR, 2021.

\bibitem[Jung et~al.(2020)Jung, Yoo, and Kang]{jung2020signed}
Jinhong Jung, Jaemin Yoo, and U~Kang.
\newblock Signed graph diffusion network, 2020.
\newblock Preprint at \url{https://arxiv.org/abs/2012.14191}.

\bibitem[Choi et~al.(2023)Choi, Hong, Park, and Cho]{choi2023gread}
Jeongwhan Choi, Seoyoung Hong, Noseong Park, and Sung-Bae Cho.
\newblock Gread: Graph neural reaction-diffusion networks.
\newblock In \emph{{International Conference on Machine Learning}}, pages
  5722--5747. PMLR, 2023.

\bibitem[Fruchterman and Reingold(1991)]{fruchterman1991graph}
Thomas~MJ Fruchterman and Edward~M Reingold.
\newblock Graph drawing by force-directed placement.
\newblock \emph{{Software: Practice and Experience}}, 21\penalty0
  (11):\penalty0 1129--1164, 1991.

\bibitem[Kamada et~al.(1989)Kamada, Kawai, et~al.]{kamada1989algorithm}
Tomihisa Kamada, Satoru Kawai, et~al.
\newblock An algorithm for drawing general undirected graphs.
\newblock \emph{{Information Processing Letters}}, 31\penalty0 (1):\penalty0
  7--15, 1989.

\bibitem[Hadany and Harel(2001)]{hadany2001multi}
Ronny Hadany and David Harel.
\newblock A multi-scale algorithm for drawing graphs nicely.
\newblock \emph{{Discrete Applied Mathematics}}, 113\penalty0 (1):\penalty0
  3--21, 2001.

\bibitem[Rahman et~al.(2020)Rahman, Sujon, and Azad]{rahman2020force2vec}
Md~Khaledur Rahman, Majedul~Haque Sujon, and Ariful Azad.
\newblock Force2vec: Parallel force-directed graph embedding.
\newblock In \emph{{2020 IEEE International Conference on Data Mining (ICDM)}},
  pages 442--451. IEEE, 2020.

\bibitem[Lotfalizadeh and Al~Hasan(2023)]{lotfalizadeh2023force}
Hamidreza Lotfalizadeh and Mohammad Al~Hasan.
\newblock Force-directed graph embedding with hops distance.
\newblock In \emph{{2023 IEEE International Conference on Big Data (BigData)}},
  pages 2946--2953. IEEE, 2023.

\bibitem[Bronstein et~al.(2021)Bronstein, Bruna, Cohen, and
  Veli{\v{c}}kovi{\'c}]{bronstein2021geometric}
Michael~M Bronstein, Joan Bruna, Taco Cohen, and Petar Veli{\v{c}}kovi{\'c}.
\newblock Geometric deep learning: Grids, groups, graphs, geodesics, and
  gauges, 2021.
\newblock Preprint available at \url{https://arxiv.org/abs/2104.13478}.

\bibitem[Blakely et~al.(2021)Blakely, Lanchantin, and Qi]{blakely2021time}
Derrick Blakely, Jack Lanchantin, and Yanjun Qi.
\newblock Time and space complexity of graph convolutional networks, 2021.
\newblock Accessed: Dec. 13, 2024. [Online]. Avail- able:
  \url{https://qdata.github.io/deep2Read//talks-mb2019/Derrick_201906_
  GCN_complexityAnalysis-writeup.pdf}.

\bibitem[Bradbury et~al.(2018)Bradbury, Frostig, Hawkins, Johnson, Leary,
  Maclaurin, Necula, Paszke, Vander{P}las, Wanderman-{M}ilne, and
  Zhang]{jax2018github}
James Bradbury, Roy Frostig, Peter Hawkins, Matthew~James Johnson, Chris Leary,
  Dougal Maclaurin, George Necula, Adam Paszke, Jake Vander{P}las, Skye
  Wanderman-{M}ilne, and Qiao Zhang.
\newblock {JAX}: composable transformations of {P}ython+{N}um{P}y programs,
  2018.
\newblock URL \url{http://github.com/google/jax}.

\bibitem[Thuerey et~al.(2021)Thuerey, Holl, Mueller, Schnell, Trost, and
  Um]{thuerey2021physics}
Nils Thuerey, Philipp Holl, Maximilian Mueller, Patrick Schnell, Felix Trost,
  and Kiwon Um.
\newblock Physics-based deep learning, 2021.
\newblock Preprint at \url{https://arxiv.org/abs/2109.05237}.

\bibitem[Zhang et~al.(2019)Zhang, He, Sra, and Jadbabaie]{zhang2019gradient}
Jingzhao Zhang, Tianxing He, Suvrit Sra, and Ali Jadbabaie.
\newblock Why gradient clipping accelerates training: A theoretical
  justification for adaptivity, 2019.
\newblock Preprint at \url{https://arxiv.org/abs/1905.11881}.

\bibitem[Kingma and Ba(2014)]{kingma2014adam}
Diederik~P Kingma and Jimmy Ba.
\newblock Adam: A method for stochastic optimization, 2014.
\newblock Preprint at \url{https://arxiv.org/abs/1412.6980}.

\bibitem[Kumar et~al.(2016)Kumar, Spezzano, Subrahmanian, and
  Faloutsos]{kumar2016edge}
Srijan Kumar, Francesca Spezzano, VS~Subrahmanian, and Christos Faloutsos.
\newblock Edge weight prediction in weighted signed networks.
\newblock In \emph{{2016 IEEE 16th International Conference on Data Mining
  (ICDM)}}, pages 221--230. IEEE, 2016.

\bibitem[Richardson et~al.(2003)Richardson, Agrawal, and
  Domingos]{richardson2003trust}
Matthew Richardson, Rakesh Agrawal, and Pedro Domingos.
\newblock Trust management for the semantic web.
\newblock In \emph{{International Semantic Web Conference}}, pages 351--368.
  Springer, 2003.

\bibitem[Ansel et~al.(2024)Ansel, Yang, He, Gimelshein, Jain, Voznesensky, Bao,
  Bell, Berard, Burovski, et~al.]{ansel2024pytorch}
Jason Ansel, Edward Yang, Horace He, Natalia Gimelshein, Animesh Jain, Michael
  Voznesensky, Bin Bao, Peter Bell, David Berard, Evgeni Burovski, et~al.
\newblock Pytorch 2: Faster machine learning through dynamic python bytecode
  transformation and graph compilation.
\newblock In \emph{{Proceedings of the 29th ACM International Conference on
  Architectural Support for Programming Languages and Operating Systems, Volume
  2}}, pages 929--947, 2024.

\end{thebibliography}




% 	\bibitem{keshet2016prediction}
% 	Keshet, Renato, Alina Maor, and George Kour.
% 	\newblock Prediction-Based, Prioritized Market-Share Insight Extraction.
% 	\newblock In {\em Advanced Data Mining and Applications (ADMA), 2016 12th International 
%                       Conference of}, pages 81--94,2016.

% \end{thebibliography}

\end{document}